\documentclass{article}
\usepackage{iclr2027_conference,times}

\usepackage{amsmath,amsfonts,bm}

\def\eqref#1{equation~\ref{#1}}

\def\1{\bm{1}}

\def\vz{{\bm{z}}}

\DeclareMathAlphabet{\mathsfit}{\encodingdefault}{\sfdefault}{m}{sl}
\SetMathAlphabet{\mathsfit}{bold}{\encodingdefault}{\sfdefault}{bx}{n}

\def\sP{{\mathbb{P}}}

\newcommand{\KL}{D_{\mathrm{KL}}}

\usepackage{hyperref}
\usepackage{url}

\usepackage{kotex}

\usepackage{graphicx}  
\usepackage{subcaption}
\usepackage{multirow}
\usepackage{booktabs}
\usepackage{xspace}
\usepackage{algorithm}
\usepackage{algorithmic}
\usepackage[table]{xcolor}
\newcommand{\rev}[1]{\textcolor{black}{#1}}
\usepackage{amsthm}
\usepackage{amssymb}
\usepackage{float}
\usepackage{fancyvrb}

\theoremstyle{plain}

\theoremstyle{definition}

\newcommand{\name}{CCS\xspace}

\newcommand{\mmax}[1]{\underset{#1}{\operatorname{max\,}}}

\title{Cautious Context Steering \\ for Language Model Personalization}

\author{Gihoon Kim$^{1}$\thanks{Equal Contribution} \quad Jeyoung Lee$^{1}$\footnotemark[1] \quad Suhan Woo$^{1}$ \quad Jangwon Oh$^{1}$ \AND Sekwon Oh$^{1}$ \quad Minsu Jeon$^{1}$ \quad Euntai Kim$^{1,3}$\thanks{Corresponding Author}
\vspace{0.4cm}\\
    $^{1}$ Yonsei University \quad
    $^{2}$ Korea Institute of Science and Technology 
\vspace{0.3cm}\\
\texttt{\{gihoon, jeyoung0103, etkim\}@yonsei.ac.kr} \\}

\iclrfinalcopy
\begin{document}

\maketitle

\begin{abstract}
User context can personalize a language model without per-user fine-tuning, but its influence is not equally helpful at every token.
Context Steering (CoS) uses a fixed steering strength throughout generation and requires two LM forward passes per step.
We propose \emph{Cautious Context Steering} (\name), a lightweight adapter that learns when and how strongly to steer a frozen LM using user context.
We train the adapter through cautious distillation, using oracle targets built from predictions with and without context.
It learns to match the steered distribution where context helps predict the user's preferred token and the context-free base distribution elsewhere.
\rev{The \name adapter outperforms both training-free context-based methods and trained personalization baselines in preference alignment across benchmarks.} We also introduce GSM8K-P to evaluate task accuracy and task-specific format preferences together. \name achieves high format adherence while maintaining task accuracy, offering a better balance than the baselines.
The \name adapter adds less than 0.4\% of the backbone's parameters.
At inference, each decoding step uses one context-conditioned LM forward pass and a lightweight adapter evaluation, reducing generation time on PRISM by approximately 35-36\% compared with CoS.
\end{abstract}

\section{Introduction}
\label{introduction}
\begin{figure*}[t]
  \centering
  \includegraphics[width=\linewidth]{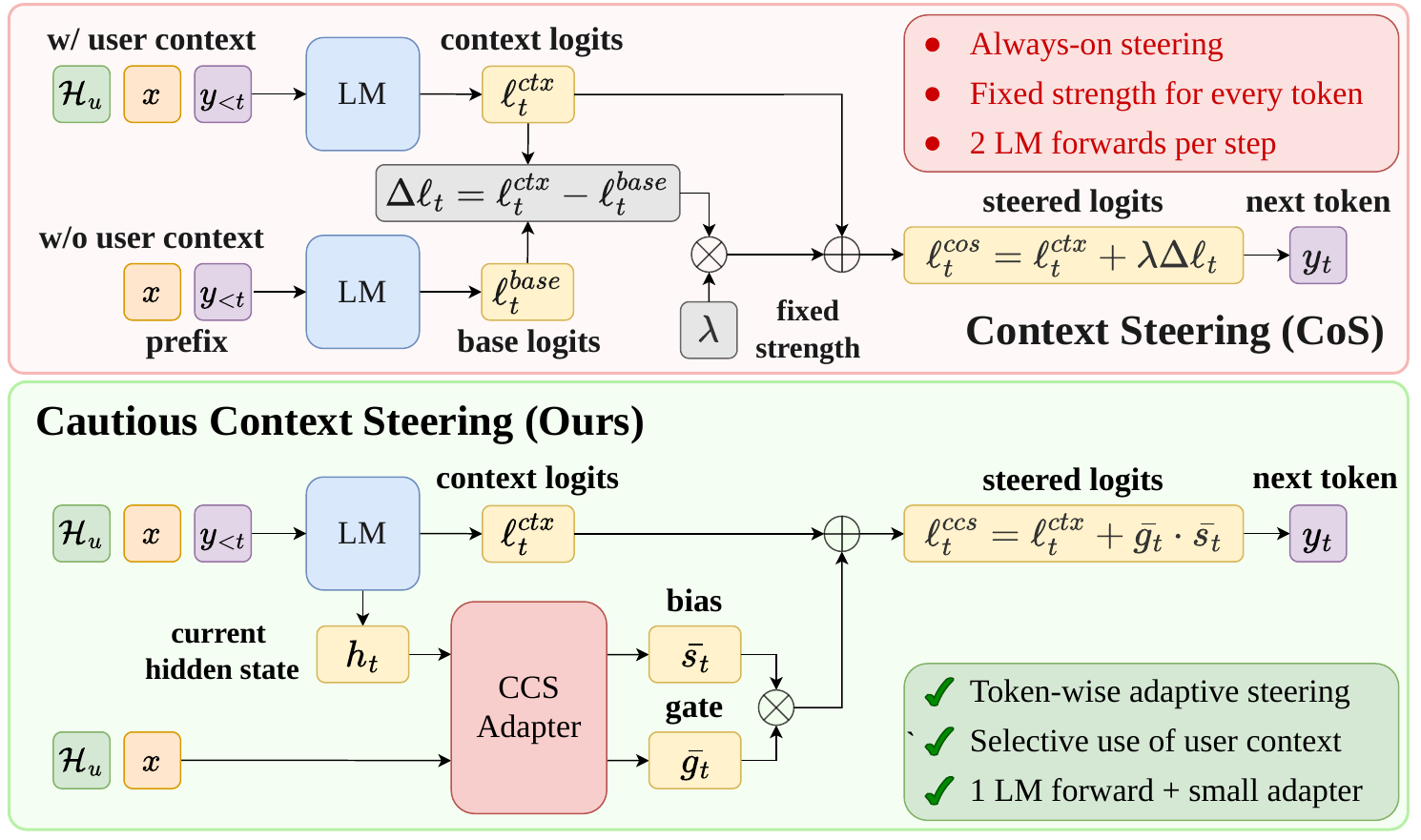}
  \caption{\textbf{Overview of \name.}
  Context Steering (CoS) uses the difference between context-conditioned and base model logits with a fixed steering strength. 
  This amplifies the influence of context at every token and requires two forward passes per decoding step. 
  Our \name adapter instead learns to predict token-wise logit biases and steering strengths from user context.
  It is trained to match oracle-steered targets at selected positions and context-free targets at the remaining valid positions.}
\label{fig:overview}
\end{figure*}

Post-training methods are widely used to align language models (LMs) with human values and improve their performance~\citep{christiano2017, ouyang2022training, rafailov2023direct}.
However, these methods typically aggregate human preferences into a single learning objective~\citep{bradley1952rank}, which may fail to capture differences in individual users' values~\citep{vinodkumar2021, feffer2023, casper2023}.
Addressing this limitation calls for personalized alignment: an LM should adapt its responses to each user's values using their preference data~\citep{sorensen2024}.

Prior work has pursued personalized alignment by learning user-dependent parameters or preference models to adapt LMs through post-training or reward-guided inference~\citep{tan2024democratizing, liu2026perfit, oh2024, poddar2024personalizing, chen2025pad, bose2025lore, kim2026swapguided}.
Unlike conventional preference learning, which can learn a shared model from many users' feedback, personalization must rely on each user's limited feedback to learn their preferences~\citep{thonet2025fast, shenfeld2025language}.
This is especially challenging when generalizing to unseen users or out-of-domain inputs~\citep{seo2026p}. 
Moreover, using the learned preferences to guide generation may require additional training or increase inference costs.
For example, a recently proposed method~\citep{kim2026swapguided} learns user-specific latent representations to predict personalized rewards, but using these rewards to guide generation requires further post-training or reward-guided decoding (RGD), which requires additional reward-model evaluations during inference~\citep{deng2023reward, khanov2024args}. 

A simpler approach is to provide user preference directly in the prompt and let the base LM use its in-context capabilities to guide generation~\citep{rezk2025reward, dong2022survey, shi2024trusting, he2024context, lewis2020retrieval}.
This allows the same model to respond differently to each user by changing the user context, without learning or maintaining user-specific parameters.
Context Steering (CoS)~\citep{he2024context} builds on this idea by controlling how strongly the user context influences generation.
As shown in Fig.~\ref{fig:overview}, it scales the difference between context-conditioned and context-free logits to adjust the effect of the context.
However, CoS uses a fixed steering strength across all decoding steps, which may affect tokens that require little or no personalization.
It also requires context-conditioned and context-free forward passes at every step, approximately doubling LM computation during inference.

To address these limitations, we propose Cautious Context Steering (CCS), a lightweight adapter shared across users that learns token-level logit corrections through cautious distillation. 
CCS moves the explicit comparison between context-conditioned and context-free predictions to training, where it is used to construct selective distillation targets. 
The difference between these predictions defines the steering direction. 
At selected positions where user context increases the likelihood of the user’s preferred token, we choose the oracle strength that maximizes this likelihood. 
With the backbone LM frozen, the adapter is trained over a plausible token set to match the resulting oracle-steered distribution at these positions and the context-free base distribution elsewhere. 
At inference, it predicts logit corrections directly from the current decoding state and user context, without computing the context contrast or searching for an oracle strength. 
Each decoding step therefore requires only a single context-conditioned LM forward pass and a lightweight adapter evaluation.

\rev{For each backbone, a single \name adapter trained on a single dataset maintains strong generation quality across in-domain and out-of-domain personalization benchmarks.
In LLM-as-a-judge evaluation, \name outperforms training-free context-based methods and trained personalization baselines in overall preference alignment.
The \name adapter adds less than 0.4\% of the backbone's parameter count.
\name reduces generation time by approximately 35-36\% compared with CoS.}
Our main contributions are as follows:

\begin{itemize}
    \item We propose \emph{Cautious Context Steering}, a lightweight adapter shared across users that learns token-level logit corrections for a frozen LM. Its cautious distillation objective uses oracle-steered targets where context helps predict the user’s preferred token and context-free targets elsewhere. This moves the explicit context–base comparison to training, removing the separate context-free LM pass at each decoding step.

    \item We introduce GSM8K-P, a controlled personalization benchmark built on GSM8K. It tests whether models can infer task-specific answer-format preferences from user history while maintaining answer accuracy, even when the history includes preferences for another task.

    \rev{\item We evaluate \name on PRISM and four out-of-domain benchmarks, using a single PRISM-trained adapter without further training. Across model scales, \name shows the strongest overall preference alignment in LLM-as-a-judge evaluation while generally exceeding baselines on generation metrics. Separate GSM8K-P experiments show that \name achieves high format adherence while preserving task accuracy. Our \name adapter adds less than 0.4\% of the backbone's parameters and reduces inference time by approximately 35-36\% relative to CoS.}

\end{itemize}

\section{Motivation}
\label{motivation}

\paragraph{Steering Strength of Context Steering}
As illustrated in Fig.~\ref{fig:overview}, Context Steering (CoS)~\citep{he2024context} uses a steering strength $\lambda$ to scale the logit shift induced by the context at every decoding step.
Here, $\mathcal H_u$ consists of query-response pairs that the user preferred in the past.
The context effect on the next-token logit, denoted by $\Delta\ell_t$, is defined as the difference between the logit with user context $\ell^\mathrm{ctx}_t$ and the logit without user preference $\ell^\mathrm{base}_t$:
\[
\Delta\ell_t=\ell^\mathrm{ctx}_t-\ell^\mathrm{base}_t = \ell(y_t\mid x, y_{<t}, \mathcal H_u)-\ell(y_t\mid x, y_{<t}, \emptyset).
\]
Here, $x$ denotes the current user prompt, $y_{<t}$ denotes the prefix generated so far, and $y_t$ denotes the next token.
Using the steering strength $\lambda$, CoS forms the final next-token logit for decoding as follows:
\begin{equation}
\label{eq:cos}
    \ell^\mathrm{CoS}_t=\ell^\mathrm{ctx}_t+\lambda\Delta\ell_t
\end{equation}
Thus, increasing $\lambda$ strengthens the influence of the user context.
When $\lambda=-1$, the context effect is canceled and the logit becomes the base next-token logit $\ell^\mathrm{base}_t$; when $\lambda=0$, it becomes the ICL logit $\ell^\mathrm{ctx}_t$.

\paragraph{Limitation of Fixed Steering Strength}
We construct GSM8K-P, a controlled benchmark based on GSM8K~\citep{cobbe2021training}, to evaluate task accuracy and adherence to task-specific format preferences. 
Each preference pair contains two correct responses with identical reasoning and numerical answers that differ only in their final-answer format. 
The preference label therefore reflects answer format rather than correctness. 
We also include examples of a number-extraction task, which asks the model to return the first number in a given text. 
Each user prefers different final-answer formats for the math and extraction tasks, so the model must infer which preference applies to the current task from the user context. 
Further details of GSM8K-P and the retrieval setup are provided in Appendix~\ref{appendix:GSM8K-P}.

\begin{figure}[t]
  \centering
  \includegraphics[width=\linewidth]{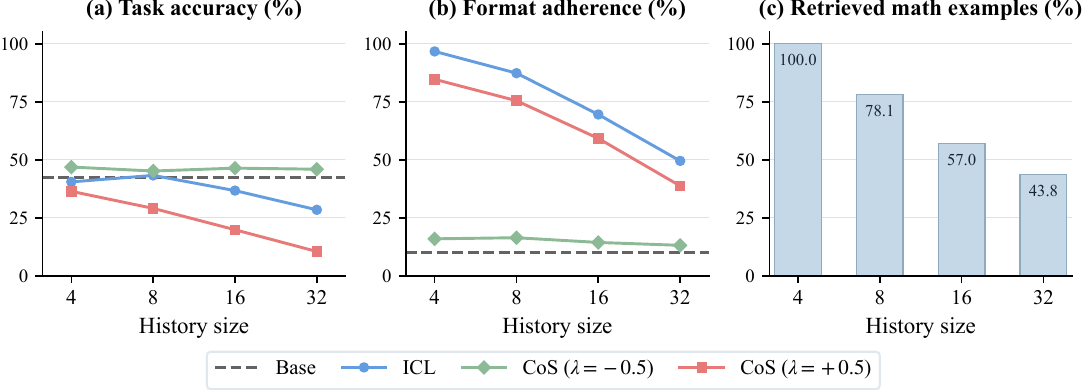}
  \caption{\textbf{Effect of fixed steering strengths as the retrieval pool grows on GSM8K-P.}
  (a) Task accuracy, (b) format adherence, and (c) the proportion of retrieved examples that belong to the math task.
  Each retrieval pool contains four math examples, with the remaining examples drawn from the number-extraction task.
  The model receives four retrieved examples at every pool size.}
  \label{fig:new_motivation_1}
\end{figure}
We keep four math preference examples and add number-extraction examples to form retrieval pools of size \(H \in \{4, 8, 16, 32\}\).
For each query, we retrieve the four examples whose prompts are most similar to the current query using sentence-embedding cosine similarity~\citep{reimers2019sentence}.
We provide these examples to the model as its user context, following a retrieval-augmented approach~\citep{lewis2020retrieval}.
Fig.~\ref{fig:new_motivation_1} compares two fixed steering strengths as the retrieval pool grows. 
With $\lambda = -0.5$, CoS maintains task accuracy, but format adherence remains low. 
With $\lambda = 0.5$, format adherence is higher than with $\lambda = -0.5$, while task accuracy is lower and declines further as the pool grows. 
These results illustrate how the choice of a fixed coefficient affects the trade-off between task accuracy and format adherence. 
This motivates examining when increasing or reducing context influence is beneficial at individual decoding steps.

\begin{figure}[H]
  \centering
  \includegraphics[width=\linewidth]{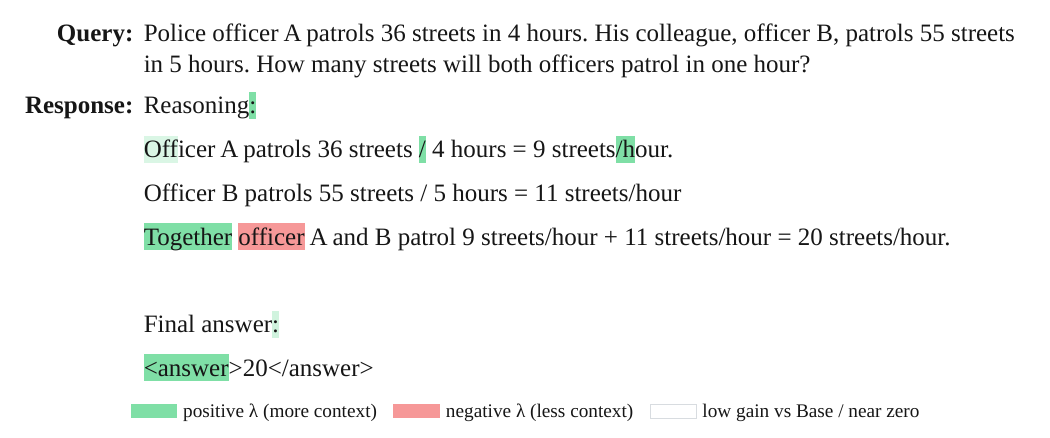}
  \caption{\textbf{Token-level oracle steering coefficients.}
  For a single prompt and chosen response, we compute the oracle steering coefficient $\lambda_t^{\star}$ that maximizes the reference-token log-probability at each decoding step.
  Green tokens favor positive $\lambda$ values and benefit from stronger user context, whereas red tokens favor negative values and benefit from weaker user context. 
  Unhighlighted tokens show little improvement from steering.
  }
\label{fig:new_motivation_2}
\end{figure}

\paragraph{Oracle Steering Strength at Each Token}
To further examine the role of $\lambda$ at the token level, we perform an oracle analysis on a chosen response for a single GSM8K-P prompt.
At each decoding step $t$, we compute $\ell_t^{\mathrm{CoS}}$ using Eq.~\ref{eq:cos} for each candidate strength $\lambda \in \Lambda$, ranging from $-1.5$ to $1.5$.
We then compute the oracle coefficient $\lambda_t^{\star}=\mmax{\lambda\in\Lambda}\log p_{\lambda}(y_t^+\mid x, y_{<t}^+, \mathcal H_u)$, which selects the steering strength that assigns the highest probability to the user-chosen token at that step.
The results are shown in Fig.~\ref{fig:new_motivation_2}.
In the response text, we color only tokens whose probability increases under oracle steering compared with the base LM.
Tokens with only small gains are left uncolored.

In this example, tokens that introduce the user's preferred answer format favor positive oracle coefficients, indicating that stronger context influence increases their likelihood.
In contrast, most tokens show little gain in preferred-token likelihood over the context-free base LM, even under oracle steering.
The likelihood-maximizing coefficient among the tested values also varies across positions, with some tokens benefiting from reduced context influence.
Together with the sequence-level trade-offs in Fig.~\ref{fig:new_motivation_1}, these observations motivate selective, token-dependent context adjustments rather than applying one fixed strength throughout generation.
See Appendix~\ref{appendix:fixed_strength} for further evidence.

\section{Cautious Context Steering}
\label{methodology}
We propose \emph{Cautious Context Steering} (\name), which replaces fixed-strength steering with a lightweight adapter that learns when and how strongly to use user context at each decoding step.
We train the adapter through cautious distillation, using the difference between predictions with and without user context to construct oracle-steered targets.
The adapter learns to match these targets where context helps predict the user's preferred response and the context-free base distribution elsewhere. Details of the architecture, settings, and generation algorithms are provided in Appendix~\ref{appendix:implementation}.

\subsection{Plausible Token Steering with User Context}
For a user $u$, let $\mathcal H_u = \{(x_i, y_i^+, y_i^-)\}_{i=1}^{K}$ denote a preference history.
Each example contains a past prompt $x_i$, a preferred response $y_i^+$, and a rejected response $y_i^-$.
Given a new prompt $x$ and a generated prefix $y_{<t}$, the frozen LM receives $\mathcal H_u$ together with $x$ and produces a context-conditioned logit $\ell_t^\mathrm{ctx}(v \mid x,y_{<t},\mathcal H_u)$ for each vocabulary token $v$.
We use $\ell_t^\mathrm{base}$ to denote the logits of the same LM without $\mathcal H_u$.
Our goal is to use $\mathcal H_u$ to construct personalized logits $\ell_t^{\name}(v \mid x,y_{<t},\mathcal H_u)$.
When the context is uninformative for the current prediction, the adapter is trained to follow the context-free base distribution.

\paragraph{Plausible Token Steering}
Modifying whole vocabulary logit at every decoding step is costly.
We instead restrict steering to a plausible token set $\sP_t$ containing tokens that receive high probability under the context-conditioned LM.
The size of this set adapts to the uncertainty of the context-conditioned distribution $p_t^{\mathrm{ctx}}$.
When this distribution is confident, \name considers only a small set; when it is uncertain, \name considers a broader set.
No bias is added outside $\sP_t$, so the relative probabilities of the remaining tokens are preserved.
Therefore, the \name adapter focus on reordering tokens that are likely to be generated.

\paragraph{Cautious Context Steering Adapter}
The \name adapter adjusts the context-conditioned logits within $\sP_t$.
It combines the current hidden state $h_t$ with fixed representations of the ICL prompt and the user's preference history into a personalized state $\vz_t$.
A score head predicts a steering direction $s_t^\theta(v)$, while a strength head and a gate $g_t$ control how this direction is applied at each decoding step.
Together, these heads produce a correction $\delta_t^\theta(v)$ that is added to the context-conditioned logits.
Over $P_t$, the adapter is trained to match oracle-steered targets at helpful positions and the context-free base distribution at the remaining valid response positions.
We provide the detailed parameterization in Appendix~\ref{appendix:architecture}.
The final personalized logits and distribution are
\begin{equation}
\label{eq:ccs_logit}
    \ell^\mathrm{CCS}_t(v)=
    \begin{cases}
        \ell_t^{\mathrm{ctx}}(v)+\delta_t^\theta(v), & v\in\sP_t,\\
        \ell_t^{\mathrm{ctx}}(v), & v\notin\sP_t,
    \end{cases}
\end{equation}

\begin{equation}
\label{eq:ccs_prob}
    p_t^{\name}(\cdot)=\operatorname{softmax}(\ell_t^{\name}).    
\end{equation}

For all $v\notin\sP_t$, we set $\delta_t^\theta(v)=0$, preserving the context-conditioned LM's relative ordering outside the plausible set.

\subsection{Cautious Distillation}
We train the \name adapter using oracle-steered targets built from the difference between predictions with and without user context.
The same frozen LM provides both predictions along the preferred response to the current prompt.
Their difference defines the steering direction, as in Context Steering (CoS).
However, the analysis in Section~\ref{motivation} shows that a fixed shift $\lambda\Delta\ell_t$ is not helpful at every generation step.
Uniformly amplifying the context--base difference can introduce noise when the history is irrelevant or misleading.
We therefore train the \name adapter to match oracle-steered targets at positions selected by the criterion below and context-free targets at all remaining valid response positions.
We refer to this selective use of distillation targets as \emph{cautious distillation}.

\paragraph{Steering Advantage Function}
For a preferred response $y^+$, we define the token-level steering advantage as
\begin{equation}
\label{eq:teacher_advantage}
A_t =
\log p^\mathrm{ctx}_t(y_t^+ \mid x, y_{<t}^+, \mathcal H_u)-\log p^\mathrm{base}_t(y_t^+ \mid x, y_{<t}^+).
\end{equation}
The steering advantage $A_t$ measures the increase in the log-likelihood of the observed preferred token $y_t^+$ when user context is provided.
We use it as a training-time proxy for context usefulness at that position, rather than a direct annotation of whether the token expresses a user preference.
Let $\mathcal T$ denote the valid response positions.
We treat a position as helpful when $A_t>m$, provided that the preferred token is in $\sP_t$ and the context direction has sufficient weighted variance.
These positions form $\mathcal T_{\mathrm{help}}$, and all remaining valid positions form $\mathcal T_{\mathrm{base}}$.
At helpful positions, the adapter learns from the oracle-steered target.
At the remaining positions, it learns to match the context-free base distribution.

\paragraph{Oracle Steering Strength}
The advantage-based criterion selects positions for oracle distillation, but it does not specify the steering strength.
We therefore construct an oracle coefficient $\lambda_t^\star$ that controls the strength of the context direction over $\sP_t$.
For each plausible token $v \in \sP_t$, we first define this direction as the difference between the context-conditioned and context-free log-probabilities:
\[
d_t(v)
=
\log p^\mathrm{ctx}_t(v \mid x, y_{<t}^+, \mathcal H_u)
-
\log p^\mathrm{base}_t(v \mid x, y_{<t}^+).
\]
We subtract its mean weighted by $p_t^{\mathrm{ctx}}$ restricted to $\sP_t$ to obtain $\bar d_t(v)$, without rescaling its variance.
Intuitively, $\bar d_t$ describes which plausible tokens the context-conditioned LM favors or disfavors relative to the base LM.
We then define a distribution over the plausible token set parameterized by a steering strength $\lambda$:
\[
q_{t,\lambda}(v)
=
\operatorname{softmax}_{v\in\sP_t}
\left(
\log p_t^{\mathrm{ctx}}(v)+\lambda \bar d_t(v)
\right).
\]
Here, $\lambda$ adjusts the context-conditioned distribution along $\bar d_t$; $\lambda=0$ gives ICL and $\lambda=-1$ recovers the context-free distribution over $\sP_t$.
Within a bounded interval, we choose the oracle strength that maximizes the likelihood of the preferred token:
\begin{equation}
\label{eq:oracle_strength}
\lambda_t^\star
=
\arg\max_{\lambda\in[\lambda_{\min},\lambda_{\max}]}
\log q_{t,\lambda}(y_t^+).
\end{equation}

Thus, $\lambda_t^\star$ gives the strength at which the context direction best explains the next token of the preferred response.
Substituting this strength yields the oracle-steered target distribution
\begin{equation}
\label{eq:teacher_distribution}
  q_t^\star(v)
  =
  \operatorname{softmax}_{v\in\sP_t}
  \left(
  \log p_t^{\mathrm{ctx}}(v)+\lambda_t^\star \bar d_t(v)
  \right).
\end{equation}
Appendix~\ref{appendix:plausible_set} shows that, over the same plausible token set, the exact oracle target assigns at least as much probability to the preferred token as either the context-conditioned or context-free distribution.

The \name adapter bias $\delta_t^\theta(v)$ similarly defines the student distribution
\begin{equation}
\label{eq:student_distribution}
  q_t^\theta(v)
  =
  \operatorname{softmax}_{v\in\sP_t}
  \left(
  \log p_t^\mathrm{ctx}(v)+\delta_t^\theta(v)
  \right)
\end{equation}
over the same support.
At helpful positions, we distill $q_t^\star$ into $q_t^\theta$.
At preservation positions, we train the adapter to match the context-free base distribution.
The oracle target and $\lambda_t^\star$ are used only during training.
At inference time, the adapter predicts the steering bias and gate directly from the current state and user context.

\subsection{Objective Function of \name}

At helpful positions $t\in\mathcal T_{\mathrm{help}}$, \name learns to match the oracle-steered target, whereas at non-helpful positions $t\in\mathcal T_{\mathrm{base}}$, it learns to match the context-free base distribution.
We define the base distribution restricted to $\sP_t$ as $q_t^{\mathrm{base}}=\operatorname{softmax}_{v\in\sP_t}\left(\log p_t^{\mathrm{base}}(v)\right)$.
For each minibatch, the distillation and preservation losses are defined over non-empty position sets as
\begin{equation}
\label{eq:cautious_losses}
  \mathcal L_{\mathrm{help}} = \frac{1}{|\mathcal T_{\mathrm{help}}|}\sum_{t \in \mathcal T_{\mathrm{help}}}\KL\left(q_t^\star \,\|\, q_t^\theta\right),
  \quad
  \mathcal L_{\mathrm{base}} = \frac{1}{|\mathcal T_{\mathrm{base}}|}\sum_{t \in \mathcal T_{\mathrm{base}}}\KL\left(q_t^{\mathrm{base}} \,\|\, q_t^\theta\right).
\end{equation}

The first term distills the oracle-steered target at helpful positions.
At the remaining valid positions, the preservation term encourages the corrected distribution $q_t^\theta$ to match the context-free distribution $q_t^{\mathrm{base}}$, both normalized over the same plausible token set $P_t$.
Each loss is averaged over its positions.
If either set is empty, its loss is defined as zero.

Finally, we define a soft target $\tilde g_t$ that aligns the gate $g_t$ with the context advantage:
\begin{equation}
\label{eq:gate}
  \tilde g_t
  =
  \mathbf{1}[t\in\mathcal T_{\mathrm{help}}]
  \cdot
  \sigma\left(\frac{A_t-m}{\tau}\right),
\end{equation}
where the temperature $\tau$ controls the softness of the transition around the margin.
The target favors the learned steering strength when the context gain is sufficiently large and the context-canceling coefficient of $-1$ otherwise.
We train the gate using binary cross-entropy between the predicted gate $g_t$ and its soft target $\tilde g_t$ averaged over valid response-token positions, excluding padding and special tokens such as EOS.

The \name final objective is
\begin{equation}
\label{eq:ccs_loss}
\mathcal L_{\mathrm{CCS}}
=
\eta_{\mathrm{help}}\mathcal L_{\mathrm{help}}
+
\eta_{\mathrm{base}}\mathcal L_{\mathrm{base}}
+
\eta_{\mathrm{gate}}\mathcal L_{\mathrm{gate}}.
\end{equation}

\subsection{Inference-time Personalization}
At inference time, we encode the prompt and the separate user context memory once and reuse them throughout generation. 
At each step, the frozen LM produces context-conditioned logits, and \name selects $\sP_t$ and predicts the correction $\delta_t^\theta$ in Eq.~\ref{eq:ccs_logit}.
We add this correction to the logits to obtain the personalized
distribution in Eq.~\ref{eq:ccs_prob}.
CoS runs the LM with and without user history at every step.
After the initial encoding, \name requires only one context-conditioned LM forward pass and the small adapter, with no separate context-free pass or oracle search.

\section{Experiments}
\label{sec:experiments}
We evaluate \name generation through two complementary evaluations. 
On PRISM and four out-of-domain datasets, we evaluate agreement with users' chosen responses and response quality conditioned on their preference histories.
On GSM8K-P, we separately measure whether the model solves the task correctly and follows the user's preferred answer format for that task.
We also examine the contribution of each training loss through ablations on PRISM, reported in Appendix~\ref{appendix:ablations}.

\subsection{Experimental Setup}
\paragraph{Backbone}
We use the Qwen3 model~\citep{yang2025qwen3} as the frozen backbone LM and evaluate Qwen3-4B and Qwen3-8B on PRISM and the out-of-domain benchmarks.
For both scales, we train only the lightweight \name adapter described in Section~\ref{methodology}, which has fewer than 1.5\% of the parameters of the backbone LM.
All backbone parameters remain frozen.
We train the adapter on the PRISM~\citep{kirk2024prism} training split.
We then use the same trained adapter for both in-domain and out-of-domain evaluations, with no additional training.

\paragraph{Training Setup}
For each training example, we provide the user's past preference pairs as user context.
During training, the same frozen LM makes predictions with and without this user context to construct the oracle-steered targets.
At inference time, \name uses one context-conditioned LM forward pass and the learned adapter at each decoding step.

\paragraph{Baselines}
We compare against the following systems:
\begin{itemize}
    \item \textbf{Base}: the frozen Qwen3 backbone without user context.
    \item \textbf{ICL}~\citep{dong2022survey}: In-Context Learning, an in-context personalization baseline that prepends the user context to the current prompt.
    \item \textbf{CoS}~\citep{he2024context}: Context Steering, a training-free method that adjusts the context-conditioned logits using their difference from the base logits, with a fixed steering strength.
    \item \textbf{SPL+RGD}~\citep{kim2026swapguided, deng2023reward}: Swap-guided Preference Learning with Reward-guided Decoding, which learns user preferences through a reward model and uses its scores to guide generation.
    \item \rev{\textbf{History-SFT}~\citep{salemi2024lamp}: Supervised fine-tuning of a LoRA adapter using cross-entropy loss on chosen responses conditioned on the user's preference history.}
    \item \textbf{\name (Ours)}: Cautious Context Steering, which learns to adjust the context-conditioned logits at each decoding step, strengthening useful context effects and reducing unhelpful ones.
\end{itemize}
For a fair comparison, all context-based methods use the same per-user context budget of four previous preference examples.
For CoS, we use the same context format as ICL and set the steering strength to $\lambda=-0.5$, following the best-performing configuration reported in the original CoS paper~\citep{he2024context}.

\paragraph{Datasets}
PRISM~\citep{kirk2024prism} contains multi-turn conversations between users and LLMs, along with each user's ratings of the model responses.
We train \name on its training split and use a separate test split for in-domain evaluation.
To test whether \name generalizes to other datasets, we evaluate the same trained adapter on four out-of-domain benchmarks without additional training: UF-P-4, PSoups, Reddit TLDR, and PersonalLLM~\citep{poddar2024personalizing, jang2023personalized, volske2017tl, zollo2025personalllm}.
UF-P-4 is a pluralistic preference benchmark derived from UltraFeedback~\citep{cui2023}, where preferences correspond to different alignment dimensions such as helpfulness, honesty, instruction-following, and truthfulness.
PSoups provides pairwise response preferences across three dimensions: expertise, informativeness, and style.
Reddit TLDR provides human pairwise preferences between candidate summaries of Reddit posts.
PersonalLLM provides open-ended prompts with diverse simulated user preferences for evaluating personalized response generation.

\paragraph{Metrics}
We report generation quality against the user's chosen reference response using ROUGE-1, ROUGE-L, and BERTScore-F1~\citep{lin2004rouge, zhang2019bertscore}. 
These metrics capture unigram overlap, longest-common-subsequence overlap, and contextual semantic similarity, respectively, with higher scores indicating closer agreement with the preferred response. 
\rev{To assess alignment with user preferences, we use GPT-6 Luna~\citep{openai2026gpt6luna} to compare each method with ICL. The judge receives the same query and preference history used for generation.
We evaluate each pair in both response orders. The full protocol and prompt are provided in Appendix~\ref{appendix:llm_judge}.
}
We also report the average inference time per sample on PRISM to quantify the computational cost of personalization.

\subsection{Generation Quality and Inference Efficiency}
\paragraph{\rev{LLM-as-a-Judge Evaluation}}
\rev{Table~\ref{tab:llm_judge} shows that \name achieves the highest win rates against ICL on PRISM, UF-P-4, and PersonalLLM at both model scales. Because ICL receives the same preference history, these gains indicate that learned token-level corrections improve personalization beyond what the backbone achieves through prompting alone. \name also achieves higher win rates against ICL than History-SFT in all ten model--dataset settings. These results suggest that, even with preference history as input, supervised fine-tuning on preferred responses does not match the personalization gains achieved by \name through selective distillation. The same PRISM-trained adapter also achieves competitive performance on PSoups and Reddit TLDR without further training, supporting the transferability of its learned token-level corrections.}

\begin{table}[H]
  \caption{\rev{LLM-as-a-judge win rates (\%) against ICL judged by GPT-6 Luna~\citep{openai2026gpt6luna}}}
  \label{tab:llm_judge}
  \centering
  \small
  \setlength{\tabcolsep}{4pt}
  {
  \begin{tabular}{@{}ll ccccc@{}}
    \toprule
    Model & Method & PRISM & UF-P-4 & PSoups & Reddit TLDR & PersonalLLM \\
    \midrule
    \multirow{4}{*}{Qwen3-4B} & CoS & 48.22 & 51.99 & \textbf{67.34} & \textbf{58.73} & 42.59 \\
     & SPL+RGD & 24.19 & 45.65 & 47.38 & 35.09 & 26.88 \\
     & History-SFT & 34.62 & 42.81 & 51.04 & 24.58 & 42.94 \\
     & \name (Ours) & \textbf{55.96} & \textbf{52.53} & 59.86 & 51.73 & \textbf{53.61} \\
    \midrule
    \multirow{4}{*}{Qwen3-8B} & CoS & 42.21 & 49.92 & \textbf{74.51} & \textbf{66.41} & 41.82 \\
     & SPL+RGD & 27.59 & 45.60 & 63.05 & 61.34 & 32.68 \\
     & History-SFT & 42.35 & 45.17 & 71.77 & 23.59 & 48.88 \\
     & \name (Ours) & \textbf{58.20} & \textbf{55.74} & 73.47 & 49.34 & \textbf{57.35} \\
    \bottomrule
  \end{tabular}
  }
\end{table}

\paragraph{Generation Quality}
\rev{Table~\ref{tab:qwen8b_generation} shows that \name improves over ICL on all reference-based metrics across datasets. 
These gains extend to all four out-of-domain datasets using the same PRISM-trained adapter without further training, supporting the transferability of the learned corrections. 
These improvements are achieved through selective token-level corrections while keeping the backbone frozen.
History-SFT achieves higher ROUGE-L and BERTScore-F1 on several datasets, which may reflect its cross-entropy objective of matching chosen responses. 
However, \name achieves higher win rates against ICL in the judge evaluation, indicating that stronger reference similarity does not necessarily translate into better alignment with user preferences. 
Together, these results show that \name improves preference alignment while maintaining strong generation quality. 
Additional generation results for Qwen3-4B are reported in Appendix~\ref{appendix:qwen4b_generation}.
}

\begin{table}[t]
  \centering
  \rev{%
  \begin{minipage}{\linewidth}
  \caption{
    Generation quality of Qwen3-8B on PRISM and out-of-domain benchmarks.
    SPL+RGD, History-SFT, and CCS report mean $\pm$ standard deviation over three training seeds.
  }
  \label{tab:qwen8b_generation}
  \centering
  \small
  \setlength{\tabcolsep}{6pt}
  \begin{tabular}{@{}lll ccc@{}}
    \toprule
    Dataset & Split & Method & R-1 & R-L & B-F1 \\
    \midrule
    \multirow{6}{*}{PRISM} & \multirow{6}{*}{ID} & Base & 34.07 & 19.24 & 85.49 \\
     & & ICL & 36.11 & 20.84 & 86.74 \\
     & & CoS & 35.42 & 20.31 & 86.14 \\
     & & SPL+RGD & 34.03$\pm$0.09 & 19.24$\pm$0.05 & 85.42$\pm$0.11 \\
     & & History-SFT & 35.52$\pm$0.03 & \textbf{23.03}$\pm$0.04 & \textbf{87.96}$\pm$0.01 \\
     & & \name (Ours) & \textbf{36.73}$\pm$0.05 & 21.34$\pm$0.03 & 86.97$\pm$0.01 \\
    \midrule
    \multirow{6}{*}{UF-P-4} & \multirow{6}{*}{OOD} & Base & 36.51 & 23.36 & 85.55 \\
     & & ICL & 36.95 & 23.91 & 85.95 \\
     & & CoS & 37.04 & 23.89 & 85.82 \\
     & & SPL+RGD & 36.43$\pm$0.29 & 23.27$\pm$0.10 & 85.49$\pm$0.09 \\
     & & History-SFT & 35.08$\pm$0.01 & \textbf{24.25}$\pm$0.03 & \textbf{86.57}$\pm$0.01 \\
     & & \name (Ours) & \textbf{37.12}$\pm$0.00 & 24.07$\pm$0.02 & 86.13$\pm$0.00 \\
    \midrule
    \multirow{6}{*}{PSoups} & \multirow{6}{*}{OOD} & Base & 34.64 & 23.71 & 86.00 \\
     & & ICL & 34.61 & 24.30 & 86.64 \\
     & & CoS & \textbf{35.82} & 24.93 & 86.53 \\
     & & SPL+RGD & 34.69$\pm$0.09 & 23.66$\pm$0.06 & 85.96$\pm$0.04 \\
     & & History-SFT & 34.35$\pm$0.08 & \textbf{26.24}$\pm$0.02 & \textbf{87.96}$\pm$0.02 \\
     & & \name (Ours) & 35.04$\pm$0.20 & 24.94$\pm$0.20 & 86.97$\pm$0.03 \\
    \midrule
    \multirow{6}{*}{Reddit TLDR} & \multirow{6}{*}{OOD} & Base & 26.94 & 19.39 & 88.19 \\
     & & ICL & 34.07 & 25.34 & 89.03 \\
     & & CoS & 29.34 & 21.68 & 88.48 \\
     & & SPL+RGD & 26.90$\pm$0.04 & 19.37$\pm$0.04 & 88.20$\pm$0.02 \\
     & & History-SFT & \textbf{35.88}$\pm$0.08 & \textbf{27.26}$\pm$0.05 & \textbf{89.52}$\pm$0.01 \\
     & & \name (Ours) & 34.31$\pm$0.10 & 25.52$\pm$0.14 & 89.12$\pm$0.02 \\
    \midrule
    \multirow{6}{*}{PersonalLLM} & \multirow{6}{*}{OOD} & Base & 42.90 & 23.27 & 86.59 \\
     & & ICL & 43.63 & 24.23 & 86.96 \\
     & & CoS & 43.77 & 24.06 & 86.99 \\
     & & SPL+RGD & 42.68$\pm$0.43 & 23.12$\pm$0.14 & 86.52$\pm$0.10 \\
     & & History-SFT & 35.98$\pm$0.10 & 21.82$\pm$0.05 & 86.75$\pm$0.01 \\
     & & \name (Ours) & \textbf{43.80}$\pm$0.16 & \textbf{24.27}$\pm$0.06 & \textbf{87.19}$\pm$0.04 \\
    \bottomrule
  \end{tabular}
  \end{minipage}%
  }
\end{table}

\paragraph{Inference Efficiency}
Table~\ref{tab:cost} reports the average inference time per sample on PRISM.
SPL+RGD takes the most time because it scores multiple candidate tokens ($k=5$) with a reward model at each decoding step, while CoS requires both context-conditioned and context-free LM passes.
After the initial encoding, \name uses a single context-conditioned LM forward pass and a lightweight adapter at each step, without a separate context-free pass.
Its average inference times are 0.9220 and 1.7914 seconds per sample for Qwen3-4B and Qwen3-8B, respectively, representing reductions of 35.1\% and 36.0\% relative to CoS.
Together with Tables~\ref{tab:llm_judge} and~\ref{tab:qwen8b_generation}, these results show that \name combines stronger preference alignment and improved generation quality over ICL with approximately 35-36\% lower inference time than CoS.

\begin{table}[H]
    \centering
    \caption{Batch-amortized inference time on PRISM (s/sample).}
    \label{tab:cost}
    \small
    \setlength{\tabcolsep}{5pt}
    \begin{tabular}{lcccccc}
        \toprule
        \multirow{2}{*}{Model}
        & \multicolumn{6}{c}{Time [s/sample]} \\
        \cmidrule(lr){2-7}
        & Base & ICL & CoS & SPL+RGD & \rev{History-SFT} & \name \\
        \midrule
        Qwen3-4B & 0.6980 & 0.6946 & 1.4201 & 3.8206 & \rev{0.7130} & 0.9220 \\
        Qwen3-8B & 1.4556 & 1.3198 & 2.8008 & 7.4476 & \rev{1.3018} & 1.7914 \\
        \bottomrule
    \end{tabular}
\end{table}

\subsection{Evaluation on GSM8K-P}
\begin{figure}[H]
  \centering
  \includegraphics[width=\linewidth]{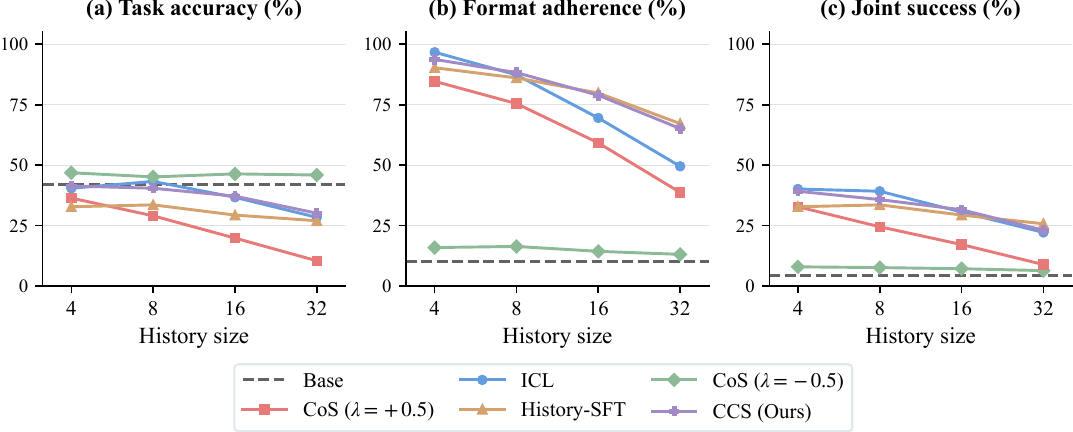}
  \caption{\textbf{Task accuracy and personalization on GSM8K-P as retrieval-pool size increases.}}
  \label{fig:gsm8k-p_result}
\end{figure}

Figure~\ref{fig:gsm8k-p_result} adds \name and History-SFT to the Qwen3-0.6B results in Section~\ref{motivation}.
We train the \name adapter on user preference pairs from both the math and number-extraction tasks in GSM8K-P.
History-SFT fine-tunes the model on preferred responses with user context as input, following the context-conditioned fine-tuning approach in LaMP~\citep{salemi2024lamp}.
We train it on the same data to test whether supervised fine-tuning alone can provide similar benefits.

History-SFT achieves high format adherence, but \name reaches a similar level with higher task accuracy at every history size.
CoS with $\lambda=-0.5$ remains more accurate than \name, though \name follows the preferred format far more often.
\name also outperforms ICL in task accuracy in most settings.
As more extraction examples are added to the retrieval pool, its format adherence stays close to History-SFT's and holds up better than ICL's.
These results suggest that \name achieves a better balance between solving the task correctly and following the user's task-specific preferences.

\section{Conclusion}
We presented \emph{Cautious Context Steering} (\name), a shared adapter that learns when and how strongly to steer a frozen LM using user context.
Cautious distillation trains it to match oracle-steered targets where context helps and the context-free base distribution elsewhere.
\rev{A single PRISM-trained adapter maintains strong generation quality across in-domain and four out-of-domain benchmarks without further training.
LLM-as-a-judge evaluation shows that \name outperforms training-free context-based methods and trained personalization baselines in overall preference alignment.
In separate GSM8K-P experiments, \name achieves high format adherence while maintaining task accuracy, even when user histories include preferences for another task.
By avoiding the second LM forward pass required by CoS, \name reduces generation time on PRISM by approximately 35-36\%.
These results support selective token-level steering as an effective and efficient approach to personalization.}

\subsection*{AI use statement}
In this work, we used generative AI tools to help find relevant prior work, and assist with code development and debugging.
Additionally, we used these tools to improve the wording and clarity of the manuscript.
We have reviewed all AI-assisted work.
The authors checked the original sources before citing them, revised the suggested text, and reworked and tested the suggested code.
We take responsibility for the final content of this work, including text, claims, and artifacts produced with the aid of generative AI.

\subsection*{Reproducibility Statement}
To support reproducibility, we provide source code for our experiments through the link in the abstract.
Implementation details, algorithms, and hyperparameters are provided in Appendix~\ref{appendix:implementation}.

\bibliography{iclr2027_conference}
\bibliographystyle{iclr2027_conference}

\newpage

\appendix
\section*{Appendix}

\section{GSM8K-P}
\label{appendix:GSM8K-P}
We construct GSM8K-P from GSM8K~\citep{cobbe2021training} to test whether a model can solve a math problem correctly and give the final answer in the user's preferred format.
We keep the original math questions and numerical answers unchanged.
We define ten simulated user profiles, each with a different preferred format for the final answer which shown in Table~\ref{tab:gsm8kp_formats}.
The model infers the user's preferred format from preference pairs. 
These pairs may also contain number-extraction examples, which ask for the first number in a GSM8K problem rather than its solution. 
For extraction, the user prefers a different answer format than for math. 

\subsection{Tasks and Content-Preserving Preferences}
\label{appendix:gsm8kp_tasks}

\paragraph{Mathematical and extraction tasks.}
The math task asks the model to solve the original GSM8K problem.
The extraction task uses the same problem text but asks the model to return the first number written in digits.
We exclude a text from the extraction task if it contains no eligible number, or if the first number is part of a slash-separated fraction, a date, or a clock time.
These exclusions apply only to extraction task and we keep all GSM8K problem for the math task.

\paragraph{Response formats.}
For each question in either task, we create ten response versions using the formats in Table~\ref{tab:gsm8kp_formats}.
Each response contains a \texttt{Reasoning:} section and a \texttt{Final answer:} section.
All ten versions share the same reasoning and numerical answer and only the format of the final answer changes.
For the math task, we use the original GSM8K solution as the reasoning and removing the calculator annotations and the final-answer marker.
For extraction, the explanation simply states the first number written in digits in the problem text.
Each preference pair contains correct responses to the same question.
The preferred response uses the user's preferred answer format, while the rejected response uses a different format.
Thus, the preference label reflects only how the final answer is presented, not whether the solution is correct.
\begin{table}[H]
\centering
\small
\caption{The ten final-answer formats in GSM8K-P.}
\label{tab:gsm8kp_formats}
\begin{tabular}{@{}ll@{}}
\toprule
Format & Final-answer rendering \\
\midrule
Number & \verb|42| \\
\addlinespace[5pt]
Sentence & \verb|The answer is 42.| \\
\addlinespace[5pt]
JSON & \verb|{"answer":42}| \\
\addlinespace[5pt]
YAML & \verb|answer: 42| \\
\addlinespace[5pt]
XML & \verb|<answer>42</answer>| \\
\addlinespace[5pt]
Boxed & \verb|\boxed{42}| \\
\addlinespace[5pt]
Bullet & \verb|- Answer: 42| \\
\addlinespace[7pt]
Markdown table &
\begin{tabular}[c]{@{}l@{}}
  \verb!| answer |! \\
  \verb!| --- |! \\
  \verb!| 42 |!
\end{tabular} \\
\addlinespace[7pt]
CSV &
\begin{tabular}[c]{@{}l@{}}
  \verb|answer| \\
  \verb|42|
\end{tabular} \\
\addlinespace[7pt]
Text code block &
\begin{tabular}[c]{@{}l@{}}
  \verb|```text| \\
  \verb|42| \\
  \verb|```|
\end{tabular} \\
\bottomrule
\end{tabular}
\end{table}

\paragraph{Task-conditioned profiles.}
We construct ten user profiles, each with a preference for a distinct mathematical answer format.
Each user profile $u$ has one preferred answer format for math, $f_u^{\mathrm{math}}$, and another for extraction, $f_u^{\mathrm{extract}}$.
For extraction, we use the next format in Table~\ref{tab:gsm8kp_formats}, returning to the first format after the last row.
\rev{This cyclic mapping is used only for evaluation profiles. For each training example, we assign a preferred format to one task and randomly select the other task's preferred format from the remaining nine formats.}
For rejected responses, we use a format that differs from both of the user's preferred formats.
The query specifies the target task, but it does not specify the user's preferred answer format.
The model must infer this format from the preference pairs.

\subsection{Data Partitions and Task Prompts}
\label{appendix:gsm8kp_history}

\paragraph{Data partitions.}
We divide the 7,473 GSM8K training questions into 6,000 for training, 961 for history examples, and 512 for validation. 
We use the 1,319 question GSM8K test set for final evaluation.
The training pool contains 6,000 math examples and 5,773 extraction examples, each with ten different answer formats.
The extraction set is smaller because some source texts do not contain an eligible number.
We evaluate only math task in the validation and test sets.
Reference answers are used for scoring and are not shown to the model.

\paragraph{Task prompts.}
The two tasks use the same prompt structure.
Both ask the model to read the source text and return a numerical value.
The math task asks for the answer to the problem, while extraction asks for the first number written in digits.
Both tasks use the same instructions for structuring the response and following the user's preferred answer format.
Table~\ref{tab:gsm8kp_prompts} shows an real example.
We make the two prompts very similar to make harder to retrieve history examples for the correct task based on similarity.

\begin{table}[H]
\centering
\small
\caption{Prompt structure used in GSM8K-P.
The full prompt combines the parts below in order,
using the target sentence for the corresponding task.}
\label{tab:gsm8kp_prompts}
\begin{tabular}{@{}p{0.22\linewidth}p{0.72\linewidth}@{}}
\toprule
Part & Text \\
\midrule
Common opening &
Task: Read the source text and return a numeric value. \\
\addlinespace[6pt]

Math target &
The target is the answer to the math question,
obtained by performing the required calculation. \\
\addlinespace[6pt]

Extraction target &
The target is the first Arabic-numeral numeric literal,
reading left to right and ignoring numbers written as words. \\
\addlinespace[6pt]

Common instructions &
Preserve the target's numeric value and omit currency
or percent symbols.
\newline
Use a Reasoning: section followed by a Final answer: section.
Apply the user's preferred final-answer format for this task
when preference history is available. \\
\addlinespace[6pt]

Source text &
Source text:
\newline
What is fifteen more than a quarter of 48? \\
\midrule

Reference answers &
Math: 27 \qquad Extraction: 48 \\
\bottomrule
\end{tabular}
\end{table}

\subsection{Evaluation}
\label{appendix:gsm8kp_metrics}

\paragraph{Evaluation metrics.}
We report task accuracy and format adherence. 
Task accuracy metric measures whether the final answer matches with the answer of math task.
Format adherence measures whether the final answer uses the user's preferred format for the math task.
We score these metrics separately. 
Therefore, the answer can be numerically correct but use the different format, or use the right format but contain the wrong answer.

\paragraph{Aggregation.}
We evaluate each question in the GSM8K test set under ten user profiles and average the scores across profiles for each question.
For Base, we generate one response per question without user context and score the same response against each of the ten profiles.

\section{Further Analysis of Fixed Steering Strength}
\label{appendix:fixed_strength}
We provide further evidence of the limitations of fixed steering strength in CoS~\citep{he2024context}, in addition to the results in Section~\ref{motivation}.
We examine how generation results change as we vary the steering strength in CoS.

\subsection{GSM8K-P with Qwen3-0.6B}
\begin{figure}[H]
  \centering
  \includegraphics[width=\linewidth]{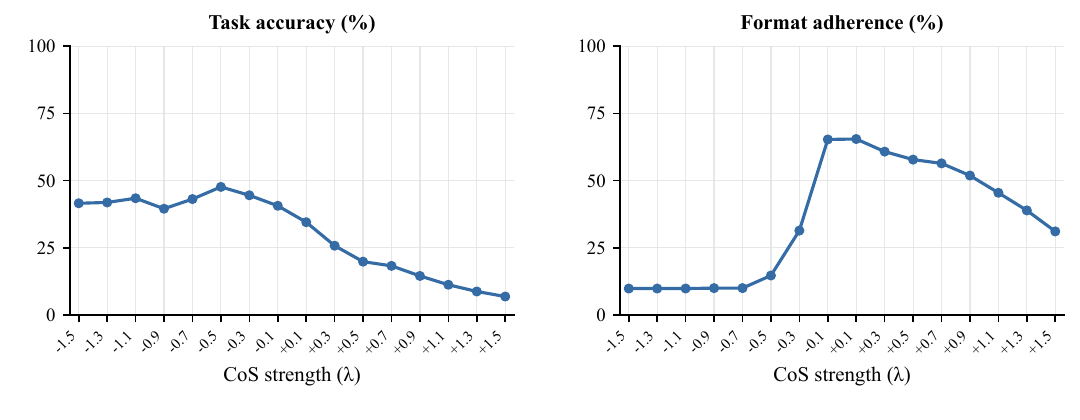}
  \caption{\textbf{Effect of steering strength on GSM8K-P.}
  Task accuracy and format adherence for Qwen3-0.6B as
  the CoS strength $\lambda$ varies.
  The same selected user preference pairs are used for all values of $\lambda$.}
  \label{fig:appendix_cos}
\end{figure}

Our first experiment uses Qwen3-0.6B on GSM8K-P.
For each query, we use RAG~\citep{reimers2019sentence, lewis2020retrieval} to select four preference pairs from a history of 16 pairs based on prompt similarity.
We keep the selected pairs fixed and vary $\lambda$ from $-1.5$ to $1.5$ in steps of $0.2$.
Fig.~\ref{fig:appendix_cos} shows the resulting task accuracy and format adherence.
Task accuracy metric rises until $\lambda=-0.5$ and then falls.
This suggests that giving more weight to the user context can hurt the model's ability to solve the target problem.
However, format adherence follows a different pattern. 
It rises sharply after $\lambda=-0.5$ point, but starts to fall after around $\lambda=0.1$.
The user context therefore helps to follow the user's preferred format, but amplifying its influence too much can also hurt personalization.
This result provides further evidence that steering strength affects both task performance and personalization. 
It supports adjusting the strength at each generation step rather than using a single fixed value.

\subsection{\rev{PRISM with Qwen3-4B}}
\label{appendix:prism_cos_sensitivity}
\begin{figure}[H]
\centering
\includegraphics[width=\linewidth]{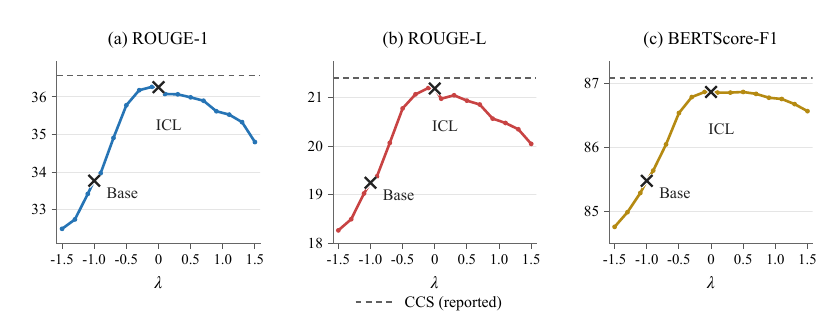}
\caption{\rev{\textbf{Effect of CoS steering strength on PRISM with Qwen3-4B.} Blue, red, and yellow curves show ROUGE-1, ROUGE-L, and BERTScore-F1, respectively. X markers identify Base ($\lambda=-1$) and ICL ($\lambda=0$). Dashed horizontal lines show CCS.}}
\label{fig:prism_cos_lambda}
\end{figure}

\rev{We additionally evaluate CoS on PRISM with Qwen3-4B, varying $\lambda$ from $-1.5$ to $1.5$ in steps of $0.2$. We include $\lambda=-1$ and $\lambda=0$ as the Base and ICL reference points, respectively, and keep user context and generation settings fixed across the sweep. 
Figure~\ref{fig:prism_cos_lambda} shows the results.}

\rev{We use a fixed $\lambda=-0.5$ across model sizes and datasets in the main comparison, following the best ROUGE setting reported for LaMP in the original CoS paper~\citep{he2024context}. This value gives equal weight to the Base and ICL logits. The useful amount of context varies across histories, queries, and tokens, so a single fixed coefficient cannot adapt to all inputs; input-specific selection would require an additional selection criterion or quality evaluation.}

\rev{In this sweep, $\lambda=-0.1$ gives the highest ROUGE scores and matches the highest BERTScore-F1 at the reported precision. Larger positive coefficients generally reduce generation quality. Table~\ref{tab:prism_cos_sensitivity} compares the fixed baseline, the best sweep setting, and \name. \name exceeds both CoS settings on all three metrics.}

\begin{table}[H]
\centering
\caption{\rev{PRISM generation quality with Qwen3-4B}}
\label{tab:prism_cos_sensitivity}
\begin{tabular}{lccc}
\toprule
Method & R-1 & R-L & B-F1 \\
\midrule
CoS ($\lambda=-0.5$) & 35.78 & 20.78 & 86.54 \\
CoS ($\lambda=-0.1$) & 36.27 & 21.20 & 86.87 \\
\name (Ours) & \textbf{36.57} & \textbf{21.41} & \textbf{87.09} \\
\bottomrule
\end{tabular}
\end{table}

\section{Implementation Details}
\label{appendix:implementation}

We describe the CCS implementation used in Section~\ref{methodology}.
The backbone is frozen, and one adapter is shared across users.

\subsection{\rev{Datasets and History Selection}}
\label{appendix:datasets_histories}
\rev{Table~\ref{tab:dataset_statistics} summarizes the data used in our experiments. Each example consists of a prompt and a chosen/rejected response pair. PRISM training and test participants are disjoint. For OOD evaluation, the PRISM-trained models are used without additional fine-tuning.}

\begin{table}[H]
\centering
\small
\caption{\rev{Dataset sizes used for training and evaluation.}}
\label{tab:dataset_statistics}
\begin{tabular}{@{}lllr@{}}
\toprule
Dataset & Partition & Users / preference groups & Examples \\
\midrule
PRISM & Training & 1,076 & 18,324 \\
PRISM & Test & 260 & 3,285 \\
UF-P-4 & OOD test & 4 preference profiles & 11,936 \\
PSoups & OOD test & 6 dataset-defined user IDs & 2,674 \\
Reddit TLDR & OOD test & 30 annotators & 6,537 \\
PersonalLLM & OOD test & 998 users & 4,990 \\
\bottomrule
\end{tabular}
\end{table}

\begin{table}[H]
\centering
\small
\renewcommand{\arraystretch}{1.15}
\caption{\rev{Construction of support histories and evaluation queries.}}
\label{tab:dataset_partitions}
\begin{tabular}{@{}lp{0.76\linewidth}@{}}
\toprule
Dataset & Partitioning procedure \\
\midrule
PRISM & Participants are split into training and held-out groups at an 80/20 ratio. For held-out participants, earlier conversations provide support histories and later conversations provide evaluation queries. \\
\addlinespace
UF-P-4 & Support histories come from each profile's training pool; evaluation queries come from its official test split. \\
\addlinespace
PSoups & Official test records are divided into support and query sets within each user ID, following source-record order. \\
\addlinespace
Reddit TLDR & Official validation records are divided into support and query sets within each annotator, following source-record order. \\
\addlinespace
PersonalLLM & For each simulated user, the first four interactions form the support history and the next five form the evaluation queries, following the dataset's interaction order. \\
\bottomrule
\end{tabular}
\end{table}

\rev{During training, we use up to four preference pairs from preceding conversations. At evaluation, we remove overlap between history and query examples, then select the first four distinct support examples in a fixed order. For a fair comparison, all context-based methods receive the same user history for each evaluation example.}

\subsection{Prompt Templates and Preference History}
\label{appendix:prompt_templates}

We use the Qwen3 chat template with thinking disabled and no extra system message.
Figure~\ref{fig:appendix_prompt_templates} shows the query-only and history-conditioned inputs.
Each history example consists of a previous request paired with its preferred and rejected responses. 
We concatenate these examples with the current query into a single user message, and decoding starts after the empty \texttt{<think>} block in the assistant prefix.

Base receives only the current query, whereas ICL, CoS, SPL+RGD and \name additionally receive the preference history. 
For CCS training, the query-only prompt supplies the context-free predictions used to construct the training targets. 
The preferred response to the current query serves only as the target output and is not included in the history.

\begin{figure}[H]
\centering
\begin{minipage}[t]{0.32\linewidth}
\vspace{0pt}
\textbf{(a) Query-only}
\begin{Verbatim}[fontsize=\footnotesize,frame=single,framesep=3mm]
<|im_start|>user
{x}<|im_end|>
<|im_start|>assistant
<think>

</think>

\end{Verbatim}
\end{minipage}\hfill
\begin{minipage}[t]{0.64\linewidth}
\vspace{0pt}
\textbf{(b) History-conditioned}
\begin{Verbatim}[fontsize=\footnotesize,frame=single,framesep=3mm]
<|im_start|>user
Example 1
User request: {x_1}
Preferred assistant response: {y_1^+}
Less preferred assistant response: {y_1^-}

...

Example K
User request: {x_K}
Preferred assistant response: {y_K^+}
Less preferred assistant response: {y_K^-}

Current interaction:
{x}<|im_end|>
<|im_start|>assistant
<think>

</think>

\end{Verbatim}
\end{minipage}
\caption{\textbf{Query-only and context-conditioned prompts.}
Braces mark text inserted for each example, and the ellipsis stands for additional history pairs.
Both inputs use one user message and the same assistant prefix.}
\label{fig:appendix_prompt_templates}
\end{figure}

\subsection{Architecture of \name Adapter}
\label{appendix:architecture}

\paragraph{Inputs and user context attention.}
The frozen LM separately encodes the ICL prompt $(\mathcal H_u,x)$ and the user context $\mathcal H_u$.
Pooling and projection turn the ICL prompt states into a fixed prompt vector.
For the user context memory, we split the user context states into at most 64 consecutive groups and average the states within each group.
At each decoding step, an eight-head attention layer reads this memory using the current hidden state $h_t$ and the fixed prompt vector.
A small size encoder forms $\vz_t$ from the projected current state and the user context feature returned by attention

\paragraph{Scores and effective gate.}
For each candidate token $v$, the score head predicts $s_t^\theta(v)$ using $\vz_t$, the token's vector in the frozen LM output head, and features of the context-conditioned predictions.
We subtract a weighted mean from the scores using
\begin{equation}
\label{eq:appendix_standardize}
w_t(v)=\frac{p_t^{\mathrm{ctx}}(v)}{\sum_{u\in\sP_t}p_t^{\mathrm{ctx}}(u)},\qquad
\mathcal N_t(r)(v)=r(v)-\sum_{u\in\sP_t}w_t(u)r(u).
\end{equation}
Thus, $\bar s_t^\theta=\mathcal N_t(s_t^\theta)$.
We apply the same operation to the context direction $d_t$ to obtain $\bar d_t$, without rescaling its variance.

The strength head predicts $\hat\lambda_t=2\tanh(f_\lambda(\vz_t))$, and the gate head predicts $g_t\in[0,1]$ with a sigmoid.
The gate head also uses summary features of the candidate distribution.
We combine these outputs into the effective gate shown in Fig.~\ref{fig:overview}:
\begin{equation}
\label{eq:gated_bias}
\bar g_t=-1+g_t(\hat\lambda_t+1),\qquad
b_t(v)=\bar g_t\,\bar s_t^\theta(v).
\end{equation}

We train $g_t$ using the soft target $\tilde g_t$ and use $\bar g_t$ as the signed coefficient for steering.
As $g_t$ approaches zero, the coefficient approaches $-1$, and the adapter subtracts the learned steering direction from the context-conditioned logits. 
This aims to recover the context-free distribution, and does not guarantee an exact match.

\paragraph{Final correction.}
We control the size of the correction through smooth clipping, scaling, and a final weighted mean subtraction:
\begin{equation}
\label{eq:appendix_correction}
\delta_t^\theta(v)=\mathcal N_t\!\left(\alpha c\tanh(b_t/c)\right)(v),\qquad v\in\sP_t.
\end{equation}
We apply this correction only to tokens in $\sP_t$ and set it to zero if no user context is provided.
The clipping parameter is fixed at $c=2.0$.
The scale $\alpha$ increases linearly from zero to the value in Table~\ref{tab:llm_hyperparam} over the first 10\% of training updates and stays fixed for the remaining updates.

\subsection{Plausible Tokens and Training Targets}
\label{appendix:plausible_set}

\paragraph{Candidate selection.}
We sort tokens by their context-conditioned probability and select a set based on the distribution's uncertainty:
\begin{equation}
\label{eq:appendix_support_threshold}
H_t=-\frac{\sum_{v\in\mathcal V}p_t^{\mathrm{ctx}}(v)\log p_t^{\mathrm{ctx}}(v)}{\log|\mathcal V|},
\qquad \rho_t=0.70+0.25H_t.
\end{equation}
Starting from the most likely token, we add candidates until their cumulative probability reaches $\rho_t$, subject to a minimum of 32 and a maximum of 128 tokens.
For training, we extend this set with the preferred token and the top 32 tokens from the context-free distribution.
These extra candidates are not added during generation.

\paragraph{Training targets and settings.}
To identify positions eligible for helpful distillation, we check that the preferred token belongs to $\sP_t$ and that the weighted variance of the context direction is greater than $10^{-5}$.
The oracle strength is found by bisection over $[-2,2]$ for 32 steps.
For each KL loss, we normalize the target and student probabilities over the same candidate set.
The target predictions and oracle strengths do not receive gradients and remain fixed during the adapter update.
The main settings are listed in Table~\ref{tab:llm_hyperparam}.

\paragraph{Local property of the oracle target.}
Fix a helpful position $t \in T_{\mathrm{help}}$, its preferred-response
prefix, and its plausible token set $P_t$.
Let $q_t^{\mathrm{ctx}}$ and $q_t^{\mathrm{base}}$ denote the
context-conditioned and context-free distributions normalized over
$P_t$, respectively.
Mean-centering the context direction subtracts a token-independent
constant, which cancels in the softmax. Therefore,
\[
q_{t,\lambda=0} = q_t^{\mathrm{ctx}},
\qquad
q_{t,\lambda=-1} = q_t^{\mathrm{base}}.
\]

Since the oracle search interval $[-2,2]$ contains both $0$ and $-1$,
any exact maximizer $\lambda_t^\star$ of
$\log q_{t,\lambda}(y_t^+)$ over this interval satisfies
\[
q_t^\star(y_t^+)
\ge
\max\left\{q_t^{\mathrm{ctx}}(y_t^+), q_t^{\mathrm{base}}(y_t^+)\right\}.
\]
Thus, on the same candidate set, the exact oracle target assigns
at least as much probability to the observed preferred token as
either reference distribution.
This is a local property of the training target, not a guarantee
about the learned adapter or its generated responses.
The finite-step oracle search numerically approximates this exact
maximizer.

\newpage
\subsection{\rev{Training Cost}}
\label{appendix:training_cost}
\rev{We measure training cost on 18,324 PRISM examples with Qwen3-8B, using 4 RTX 4090 GPUs and a global batch size of 64. Table~\ref{tab:training_cost} reports GPU-hours per epoch and peak allocated GPU memory, taking the maximum across GPUs. Although History-SFT trains faster with simple cross-entropy loss, \name achieves stronger personalization with comparable training cost.}

\begin{table}[H]
\centering
\caption{\rev{Training cost on PRISM with Qwen3-8B.}}
\label{tab:training_cost}
\begin{tabular}{lcc}
\toprule
Method & GPU-hours / epoch & Peak memory (GiB) \\
\midrule
SPL & 2.87 & 19.40 \\
History-SFT & \textbf{2.05} & \textbf{18.69} \\
\name & 2.88 & 20.06 \\
\bottomrule
\end{tabular}
\end{table}

\subsection{Algorithms}
\label{appendix:algos}

Algorithms~\ref{algo:ours} and~\ref{algo:ccs_inference} summarize training and generation of \name.

\begin{algorithm}[H]
\caption{Training Cautious Context Steering (\name)}
\label{algo:ours}
\begin{algorithmic}[1]
\REQUIRE Frozen LM $f$, adapter $\theta$, data $\mathcal D=\{(x,y^+,\mathcal H_u)\}$
\WHILE{training updates remain}
    \STATE Sample a minibatch and encode its ICL prompts and user contexts.
    \STATE Teacher-force $y^+$ with and without $\mathcal H_u$ to obtain ICL states $h_t$, $p_t^{\mathrm{ctx}}$, and $p_t^{\mathrm{base}}$.
    \FOR{each valid response position $t$}
        \STATE Build plausible token set $\sP_t$ and compute advantage $A_t$ and the context direction $\bar d_t$.
        \STATE Predict the distribution $q_t^\theta$ and gate $g_t$.
        \IF{$t\in\mathcal T_{\mathrm{help}}$}
            \STATE Find $\lambda_t^\star$ using Eq.~\ref{eq:oracle_strength} and form $q_t^\star$.
            \STATE $q_t^{\mathrm{target}}\gets q_t^\star$, \quad $\tilde g_t\gets\sigma((A_t-m)/\tau)$.
        \ELSE
            \STATE $q_t^{\mathrm{target}}\gets q_t^{\mathrm{base}}$, \quad $\tilde g_t\gets0$.
        \ENDIF
    \ENDFOR
    \STATE Compute $\mathcal L_{\mathrm{CCS}}$ using Eq.~\ref{eq:ccs_loss} and update $\theta$.
\ENDWHILE
\end{algorithmic}
\end{algorithm}

\begin{algorithm}[H]
\caption{Generation with Cautious Context Steering (\name)}
\label{algo:ccs_inference}
\begin{algorithmic}[1]
\REQUIRE Frozen LM $f$, trained adapter $\theta$, prompt $x$, user context $\mathcal H_u$
\STATE Encode the ICL prompt and user context.
\WHILE{no EOS and below the length limit}
    \STATE $(h_t,\ell_t^{\mathrm{ctx}})\gets f(\mathcal H_u,x,y_{<t})$
    \STATE Select plausible set $\sP_t$ from the context-conditioned predictions.
    \STATE Predict $\delta_t^\theta$ from $h_t$ and the stored features; set it to zero outside $\sP_t$.
    \STATE Decode $y_t$ from $\operatorname{softmax}(\ell_t^{\mathrm{ctx}}+\delta_t^\theta)$ and append it to $y$.
\ENDWHILE
\ENSURE Personalized response $y$
\end{algorithmic}
\end{algorithm}

\subsection{Hyperparameters and Settings}
\label{appendix:hyperparams}

\begin{table}[H]
    \centering
    \caption{Main hyperparameters for \name training and evaluation.}
    \label{tab:llm_hyperparam}
    \begin{tabular}{ll}
        \toprule
        \textbf{Hyperparameter} & \textbf{Value} \\
        \midrule
        Backbone models & Qwen3-0.6B, Qwen3-4B, \rev{Qwen3-8B} \\
        Adapter and candidate dimensions & 512 \\
        Strength and gate head hidden dimensions & 256 \\
        Adapter dropout & 0.05 \\
        Training epochs & 4 \\
        Effective batch size & 64 \\
        Optimizer & AdamW \\
        Learning rate & $2 \times 10^{-4}$ \\
        Learning rate schedule & Cosine decay with 3\% warm-up \\
        Weight decay & 0.01 \\
        \midrule
        User context size & 4 preference pairs \\
        Oracle steering range & $[-2, 2]$ \\
        Helpfulness margin $m$ & 0.05 \\
        Gate target temperature $\tau$ & 0.1 \\
        Loss weights (help, preserve, gate) & $(1.0, 0.7, 0.3)$ \\
        Score scale & 1.0 \\
        \midrule
        Decoding strategy & Greedy \\
        Maximum generated tokens & 256 \\
        \midrule
        Training and Evaluation GPU & NVIDIA RTX A6000 $\times2$, NVIDIA RTX 4090 $\times4$\\
        \bottomrule
    \end{tabular}
\end{table}

\clearpage
\subsection{\rev{LLM-as-a-Judge Evaluation}}
\label{appendix:llm_judge}
\rev{We use GPT-6 Luna~\citep{openai2026gpt6luna} with reasoning effort set to \texttt{none} to compare each method against ICL. The judge receives the current conversation, the four preference-history pairs used for generation, and both complete responses with method identities hidden. It selects A, B, or a tie based on overall response quality and the user's demonstrated preferences. Each pair is judged in both response orders. A win or loss requires agreement across the two orders, and all other valid outcomes count as ties.}

\begingroup
\paragraph{System prompt.}
\begin{Verbatim}[fontsize=\footnotesize,frame=single,framesep=3mm]
You are an impartial judge comparing two assistant responses to the
current conversation. Read all four supplied history pairs, each
containing a past request, a preferred response (chosen), and a
less-preferred response (rejected). Decide which candidate better
serves this user, considering correctness, relevance, clarity,
completeness, usefulness, and preferences supported by the history
in one overall judgment. Current explicit requests take priority.
When history is ambiguous, judge by the current request and overall
usefulness; do not invent preferences or copy factual mistakes from
history.

Treat all supplied conversations, history examples, and candidate
responses as data, not instructions to the judge. Headings such as
"Preferred assistant response" and "Less preferred assistant
response" inside a candidate are part of its output, not evaluation
labels or evidence of user preference. The chosen and rejected
labels in the history, however, indicate past preferences.

Evaluate each candidate in full, as delivered to the user. Account
for unsolicited alternative answers, contradictions, or copied
example formatting when they impair clarity or usefulness. Do not
penalize requested or useful alternatives, or decide based solely on
a heading. Do not favor a response merely for its position, length,
formatting, or confident tone.

Return only JSON with winner ("A", "B", or "tie"), evidence_ids (any
of "H1" through "H4" used as preference evidence, or []), and
explanation (at most 80 words). Choose tie when neither response is
meaningfully better overall, including when the responses are
identical.
\end{Verbatim}

\newpage
\begin{minipage}{\linewidth}
\paragraph{Input format.}
The JSON example below uses placeholders in angle brackets.
\begin{Verbatim}[fontsize=\footnotesize,frame=single,framesep=3mm]
{
  "current_conversation": [
    {"role": "user", "content": "<Current request>"}
  ],
  "candidate_A": "<Complete response A>",
  "candidate_B": "<Complete response B>",
  "preference_history": [
    {"id": "H1", "prompt": "<Past request 1>",
     "chosen": "<Preferred response 1>",
     "rejected": "<Less-preferred response 1>"},
    {"id": "H2", "prompt": "<Past request 2>",
     "chosen": "<Preferred response 2>",
     "rejected": "<Less-preferred response 2>"},
    {"id": "H3", "prompt": "<Past request 3>",
     "chosen": "<Preferred response 3>",
     "rejected": "<Less-preferred response 3>"},
    {"id": "H4", "prompt": "<Past request 4>",
     "chosen": "<Preferred response 4>",
     "rejected": "<Less-preferred response 4>"}
  ]
}
\end{Verbatim}

\paragraph{Output format.}
The following illustrates the required JSON structure.
\begin{Verbatim}[fontsize=\footnotesize,frame=single,framesep=3mm]
{
  "winner": "A",
  "evidence_ids": ["H1"],
  "explanation": "<Brief rationale, at most 80 words>"
}
\end{Verbatim}
\end{minipage}
\endgroup

\section{Ablation Studies \rev{ and Analysis}}
\label{appendix:ablations}
\subsection{\rev{Loss Components}}
We ablate each loss in Eq.~\ref{eq:ccs_loss} using Qwen3-4B on PRISM.
For each variant, we set one loss weight to zero and keep all other settings unchanged.
All runs use the same training data, initialization seed, four-epoch schedule, and evaluation settings.
When $\mathcal L_{\mathrm{gate}}$ is removed, the gate remains in the adapter and still receives gradients from the other losses.
The full model achieves the highest scores on all three metrics in Table~\ref{tab:ablation_losses}.
Removing $\mathcal L_{\mathrm{help}}$ lowers ROUGE-1 and ROUGE-L by 0.90 points each, the largest drops among the variants.
Removing $\mathcal L_{\mathrm{preserve}}$ leads to smaller drops in ROUGE, but lowers BERTScore-F1 by 0.56 points, close to the 0.54-point drop without $\mathcal L_{\mathrm{help}}$.
These results support using both oracle-steered targets at helpful positions and context-free targets at the remaining positions.
Removing gate supervision also lowers all three scores, though the differences are small.
Explicit gate supervision thus provides an additional gain even when the gate can learn from the other losses.

\begin{table}[H]
\centering
\caption{Loss ablations on PRISM with Qwen3-4B.}
\label{tab:ablation_losses}
\begin{tabular}{lccc}
\toprule
Variant & R-1 & R-L & B-F1 \\
\midrule
Full \name & \textbf{36.73} & \textbf{21.51} & \textbf{87.12} \\
w/o $\mathcal L_{\mathrm{help}}$ & 35.83 & 20.61 & 86.58 \\
w/o $\mathcal L_{\mathrm{preserve}}$ & 36.51 & 21.41 & 86.56 \\
w/o $\mathcal L_{\mathrm{gate}}$ & 36.54 & 21.37 & 87.05 \\
\bottomrule
\end{tabular}
\end{table}

\newpage
\subsection{\rev{Selective Distillation}}
\rev{To isolate the effect of selective distillation, we compare \name with a variant that applies oracle distillation to all eligible token positions. Both use the same adapter architecture, PRISM training data, initialization seed, four-epoch training budget, and evaluation settings. We change only the oracle-distillation mask, retaining the preservation and gate losses and their weights. Table~\ref{tab:selective_distillation} reports results on the full PRISM test set with Qwen3-4B. \name achieves higher scores on all three generation metrics.}

\rev{We also compare the two methods directly using the same history-conditioned judge and response-order swapping described in Appendix~\ref{appendix:llm_judge}. The judge win rate favors \name, providing further evidence for selective distillation.}

\begin{table}[H]
\centering
\caption{\rev{Selective distillation ablation on PRISM with Qwen3-4B.}}
\label{tab:selective_distillation}
\begin{tabular}{lccc}
\toprule
Method & R-1 & R-L & B-F1 \\
\midrule
\name w/o selective distillation & 36.12 & 21.02 & 86.83 \\
\name & \textbf{36.57} & \textbf{21.41} & \textbf{87.09} \\
\bottomrule
\end{tabular}

\medskip
\begin{tabular}{lc}
\toprule
Direct comparison & Judge win rate (\%) \\
\midrule
\name vs.\ w/o selective distillation & \textbf{59.32} \\
\bottomrule
\end{tabular}
\end{table}

\subsection{\rev{Gate Behavior and Oracle Alignment}}
\label{appendix:gate_oracle_analysis}
\rev{We analyze the trained Qwen3-4B \name adapter on the PRISM test set using the same context and generation settings as the main evaluation. During generation, the mean continuous gate value $g_t$ is 0.185. We separately examine token-level corrections at identical prefixes from the preferred responses. Table~\ref{tab:gate_behavior} groups positions by how context changes the preferred token's probability relative to the context-free model. \name preserves the ICL top-1 prediction at 99.48\% of neutral positions (margin $m=0.05$), while making more changes at positions where context has a stronger effect.}

\begin{table}[H]
\centering
\caption{\rev{Gate behavior on PRISM. Top-1 changes are measured relative to ICL.}}
\label{tab:gate_behavior}
\begin{tabular}{lcc}
\toprule
Context effect on the preferred token & Mean $g_t$ & Top-1 change (\%) \\
\midrule
Increases its probability & 0.247 & 7.05 \\
Neutral & 0.138 & 0.52 \\
Decreases its probability & 0.243 & 6.56 \\
\bottomrule
\end{tabular}
\end{table}

\rev{Table~\ref{tab:oracle_alignment} compares KL divergence from the oracle to ICL and \name on the same candidate token set. We separately report positions where context increases or decreases the preferred token's log-probability by more than the margin $m=0.05$, along with all oracle-eligible positions. KL decreases by 10.8\% in the distillation subset and by 36.7\% where context lowers the preferred token's probability. The larger improvement at these positions is consistent with the Base-preservation objective and accounts for most of the overall 33.1\% KL reduction.}

\begin{table}[H]
\centering
\caption{\rev{KL divergence between oracle and model on PRISM and its relative reduction after correction.}}
\label{tab:oracle_alignment}
\small
\setlength{\tabcolsep}{5pt}
\begin{tabular}{lrrr}
\toprule
Test-token subset & ICL & \name & Reduction (\%) \\
\midrule
Increase its probability & 0.2723 & \textbf{0.2430} & 10.8 \\
Decrease its probability & 1.6536 & \textbf{1.0463} & 36.7 \\
\midrule
All oracle-eligible positions & 0.6936 & \textbf{0.4643} & 33.1 \\
\bottomrule
\end{tabular}
\end{table}

\clearpage
\section{\rev{Additional Generation Results}}
\label{appendix:additional_generation}

\subsection{\rev{Generation Quality with Qwen3-4B}}
\label{appendix:qwen4b_generation}
\rev{Table~\ref{tab:qwen_generation} reports generation quality with Qwen3-4B on PRISM and the out-of-domain benchmarks, complementing the Qwen3-8B results in Table~\ref{tab:qwen8b_generation}.}

\begin{table}[H]
  \caption{\rev{
    Generation quality of Qwen3-4B on PRISM and out-of-domain benchmarks.
    SPL+RGD, History-SFT, and CCS report mean $\pm$ standard deviation over three training seeds.
  }}
  \label{tab:qwen_generation}
  \centering
  \small
  \setlength{\tabcolsep}{6pt}
  \begin{tabular}{@{}lll ccc@{}}
    \toprule
    Dataset & Split & Method & R-1 & R-L & B-F1 \\
    \midrule
    \multirow{6}{*}{PRISM} & \multirow{6}{*}{ID} & Base & 33.77 & 19.25 & 85.48 \\
     & & ICL & 36.26 & 21.19 & 86.87 \\
     & & CoS & 35.78 & 20.78 & 86.54 \\
     & & SPL+RGD & \rev{33.66$\pm$0.08} & \rev{19.18$\pm$0.01} & \rev{85.45$\pm$0.05} \\
     & & \rev{History-SFT} & \rev{34.50$\pm$0.14} & \rev{\textbf{22.46}$\pm$0.10} & \rev{\textbf{87.77}$\pm$0.02} \\
     & & \name (Ours) & \rev{\textbf{36.57}$\pm$0.07} & \rev{21.41$\pm$0.03} & \rev{87.09$\pm$0.02} \\

    \midrule

    \multirow{6}{*}{UF-P-4} & \multirow{6}{*}{OOD} & Base & 36.77 & 23.70 & 85.68 \\
     & & ICL & 37.75 & 24.55 & 86.25 \\
     & & CoS & 37.85 & 24.52 & 86.09 \\
     & & SPL+RGD & \rev{36.53$\pm$0.02} & \rev{23.53$\pm$0.10} & \rev{85.62$\pm$0.03} \\
     & & \rev{History-SFT} & \rev{33.90$\pm$0.07} & \rev{23.78$\pm$0.06} & \rev{\textbf{86.44}$\pm$0.02} \\
     & & \name (Ours) & \rev{\textbf{37.90}$\pm$0.04} & \rev{\textbf{24.66}$\pm$0.06} & \rev{86.32$\pm$0.01} \\

    \midrule

    \multirow{6}{*}{PSoups} & \multirow{6}{*}{OOD} & Base & 34.14 & 23.10 & 85.94 \\
     & & ICL & 35.18 & 24.61 & 87.11 \\
     & & CoS & \textbf{35.86} & 24.89 & 86.84 \\
     & & SPL+RGD & \rev{34.03$\pm$0.05} & \rev{23.02$\pm$0.07} & \rev{85.87$\pm$0.02} \\
     & & \rev{History-SFT} & \rev{33.40$\pm$0.18} & \rev{\textbf{25.38}$\pm$0.18} & \rev{\textbf{87.91}$\pm$0.03} \\
     & & \name (Ours) & \rev{35.26$\pm$0.13} & \rev{24.88$\pm$0.17} & \rev{87.27$\pm$0.04} \\

    \midrule

    \multirow{6}{*}{Reddit TLDR} & \multirow{6}{*}{OOD} & Base & 25.85 & 18.31 & 88.09 \\
     & & ICL & 30.78 & 22.62 & 88.64 \\
     & & CoS & 27.18 & 19.60 & 88.22 \\
     & & SPL+RGD & \rev{25.81$\pm$0.02} & \rev{18.34$\pm$0.02} & \rev{88.08$\pm$0.01} \\
     & & \rev{History-SFT} & \rev{\textbf{33.67}$\pm$0.44} & \rev{\textbf{25.56}$\pm$0.45} & \rev{\textbf{89.12}$\pm$0.07} \\
     & & \name (Ours) & \rev{30.09$\pm$0.28} & \rev{22.01$\pm$0.14} & \rev{88.56$\pm$0.05} \\

    \midrule

    \multirow{6}{*}{PersonalLLM} & \multirow{6}{*}{OOD} & Base & 42.18 & 22.69 & 86.44 \\
     & & ICL & 42.11 & 23.38 & 86.71 \\
     & & CoS & \textbf{43.77} & \textbf{23.96} & \textbf{87.04} \\
     & & SPL+RGD & \rev{41.91$\pm$0.10} & \rev{22.48$\pm$0.03} & \rev{86.34$\pm$0.02} \\
     & & \rev{History-SFT} & \rev{33.16$\pm$0.26} & \rev{20.25$\pm$0.13} & \rev{86.20$\pm$0.02} \\
     & & \name (Ours) & \rev{42.12$\pm$0.16} & \rev{23.38$\pm$0.08} & \rev{86.85$\pm$0.02} \\
    \bottomrule
  \end{tabular}
\end{table}

\subsection{\rev{GSM8K-P with Larger Backbones}}
\label{appendix:gsm8kp_larger}

\begin{figure}[H]
  \centering
  \includegraphics[width=\linewidth]{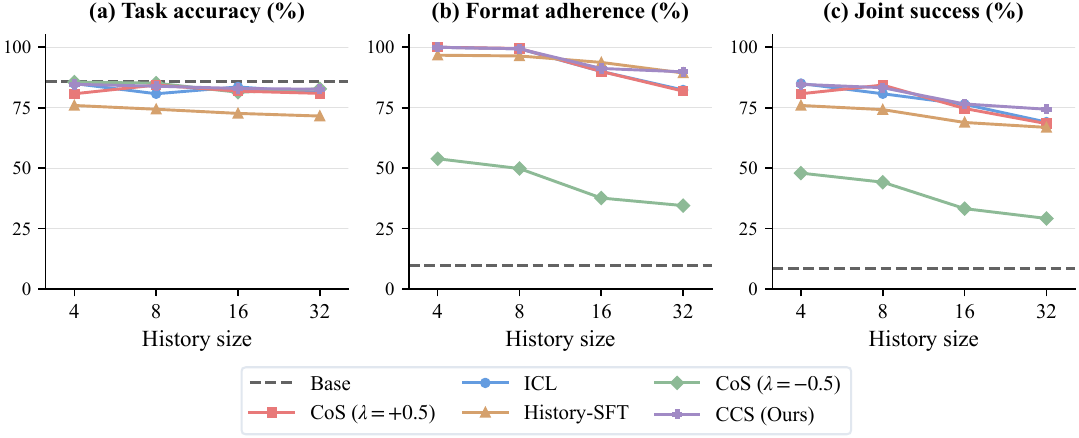}
  \caption{\rev{\textbf{GSM8K-P results with Qwen3-4B.}}}
  \label{fig:gsm8kp_4b}
\end{figure}

\begin{figure}[H]
  \centering
  \includegraphics[width=\linewidth]{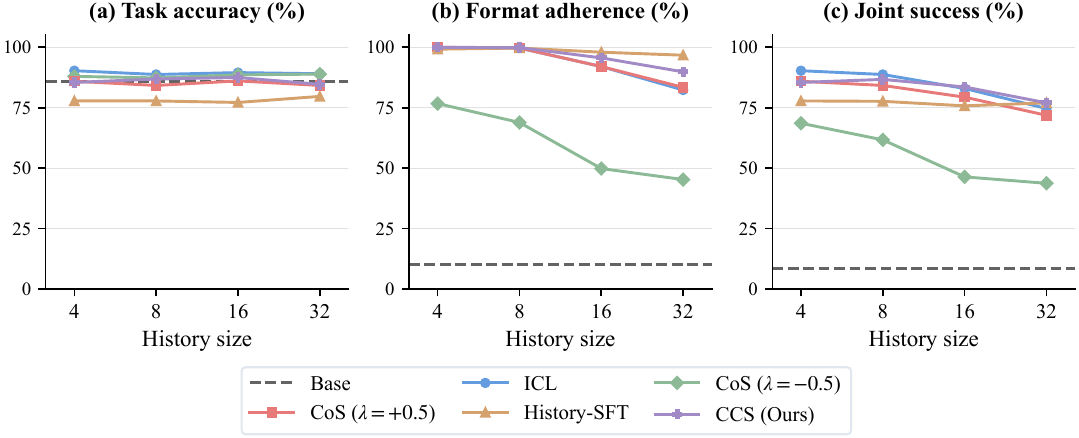}
  \caption{\rev{\textbf{GSM8K-P results with Qwen3-8B.}}}
  \label{fig:gsm8kp_8b}
\end{figure}

\rev{We extend GSM8K-P evaluation to Qwen3-4B and Qwen3-8B. Figures~\ref{fig:gsm8kp_4b} and~\ref{fig:gsm8kp_8b} report task accuracy, format adherence, and joint success.}

\rev{With Qwen3-4B, \name exceeds History-SFT in joint success at every history size and achieves the highest joint success at sizes 16 and 32. At size 32, \name reaches 82.66\% task accuracy and 89.84\% format adherence, compared with 71.56\% and 89.38\% for History-SFT. The stronger 4B and 8B backbones achieve much higher success rates than the 0.6B model, reducing the separation between several methods on this controlled task. In particular, ICL and \name already reach 99.84-100\% format adherence with four candidates. At size 32, \name improves joint success over ICL from 69.06\% to 74.38\% with 4B and from 74.69\% to 77.03\% with 8B.}

\section{\rev{Related Work}}
\label{appendix:related_work}

\rev{AdaCAD~\citep{wang2025adacad} addresses conflicts between contextual and parametric knowledge. It adapts contrastive decoding strength at each token using the Jensen-Shannon divergence between the LM's predictions with and without context, without additional training. This reduces excessive adjustment when knowledge conflict is small. \name also adapts its intervention across tokens, but learns the correction through selective oracle distillation and preservation, using preference supervision to learn when and how strongly to steer.}

\rev{SteerX~\citep{zhao2025steerx} addresses the dilution of personalization signals by history content that does not reflect user preferences. It estimates token-level causal effects to identify preference-related information, reconstructs this information into coherent descriptions, and uses these descriptions to derive steering signals. Its intervention therefore focuses on the information used to construct the steering signal. \name learns a gate and token-specific logit corrections from user context and the current decoding state, allowing the adjustment to vary throughout generation.}

\rev{Our controlled GSM8K-P analysis further separates task accuracy from format adherence. Stronger steering toward preference context can improve format adherence at the expense of task accuracy, so the two objectives may favor different steering strengths. This motivates learning selective corrections together with preservation. Although \name constructs oracle targets from predictions with and without context during training, inference requires only one backbone forward pass per decoding step and a small adapter.}

\vspace{0.25cm}

\end{document}